\documentclass[conference]{IEEEtran}

\usepackage{booktabs}
\usepackage{multirow}
\usepackage{amsmath}
\usepackage[ruled,vlined,linesnumbered,noend]{algorithm2e}
\usepackage{subcaption}
\usepackage{enumitem}
\usepackage{xurl}
\usepackage{balance}
\usepackage{makecell}
\usepackage{placeins}
\usepackage{hyperref}
\usepackage{amssymb}
\usepackage{tikz}
\usetikzlibrary{arrows.meta,positioning,fit}
\usetikzlibrary{arrows.meta,calc,backgrounds}

\newcommand{\posid}{\textsc{POSID}}
\newcommand{\incitext}{\textsc{InciText}}

\title{\posid{}: Target-Task-Unsupervised Person-Attribute Extraction from Heterogeneous Public-Safety Text}

\author{\IEEEauthorblockN{KMA Solaiman}
\IEEEauthorblockA{
University of Maryland, Baltimore County, Baltimore, MD\\
ksolaima@umbc.edu}}

\begin{document}
\maketitle
\begin{abstract}
Person descriptions in public-safety text can be compact and regular, or
distributed across heterogeneous narratives in which evidence boundaries and
person ownership are ambiguous. We present \posid{}, a
target-task-unsupervised framework that decomposes person-attribute extraction
into candidate localization, provenance-controlled extraction scope,
structured recovery and traversal, and person association, without supervised
training or fine-tuning on the target task. We evaluate six localization
configurations under two traversal policies across three real-use cohorts and
two controlled synthetic diagnostics. On structurally regular descriptions,
the strongest configurations reach 1.000/.972 Attr-Only (AO)/Attr-Value (AV)
$F_1$ on Compact500. Across the real-use cohorts, every RE-containing
configuration benefits from conservative early stopping, while semantic-only
retrieval can benefit from continued traversal when localization is less
constrained. On the heterogeneous multi-person Challenge14 cohort, the best
automatic AO and AV results are .500 and .382; replacing automatic candidates
with verified evidence raises the frozen extractor to .522/.410 under early
stopping, showing that localization is one of the material bottlenecks.
We further introduce provenance-aware scoring for collective and
underspecified evidence and a separate person-aware diagnostic for ownership
and fragmentation. Together, the results establish a measurable operating
regime for inspectable person-attribute extraction and distinguish errors due
to localization, traversal, structured recovery, and attribution.

\noindent Code: \url{https://github.com/ksolaiman/hart-posid}.
\end{abstract}

\section{Introduction}
\label{sec:introduction}

Structured descriptions of people are frequently embedded in free-form
public-safety text. Incident reports, dispatch records, press releases, and
missing-person notices may describe gender, race, height, clothing, and other
appearance attributes alongside event details, locations, witnesses, and
multiple interacting people. Recovering these descriptions is therefore not
only a recognition problem. A system must determine where relevant evidence
occurs, decide how much of the localized text belongs to the attribute
description, recover a structured representation, and associate that evidence
with the person or group ownership supported by the source. 
This setting was motivated by discussions with investigators at the West Lafayette Police Department (WLPD), who described the practical need to automatically recover person attributes from narrative reports before downstream investigative tasks such as video review. The operational setting also motivated a lightweight design suitable for modest computing environments, with inspectable structured output rather than dependence on substantial task-specific training or specialized compute infrastructure.

Meeting this need across heterogeneous reports is difficult because document
structure can vary substantially. Compact descriptions may place demographic
information and several garments in a single span, whereas longer narratives
may distribute evidence across sentences, introduce several people or roles,
and interleave person descriptions with unrelated event text. Clothing presents
an additional challenge because both item names and descriptors are
open-vocabulary, while multi-person narratives can express attributes
collectively or without specifying which member of a group owns a particular
description.

Public-safety information extraction has long combined lexical, rule-based,
and statistical NLP components to recover structured information from police
and crime-related narratives~\cite{iriberri2007crime,ku2008crimeie,skorzewski2022criminal}.
More recent work has reduced task-specific supervision through zero-shot
slot filling and entailment-based extraction~\cite{du2021qasf,sainz2021entailment}, unified or guideline-conditioned
information extraction~\cite{lu2022uie,sainz2024gollie}, and direct structured extraction with
large language models~\cite{li2024structured,shamsabadi2024llmie}.
Recent public-safety systems have also applied fine-tuned or generative
language models to police-report information extraction
~\cite{xing2024police,liu2025tips}, while recent missing-person systems use
schema-guided and multi-LLM pipelines
~\cite{castillo2026parser,castillo2026guardian}. 
This work studies a complementary operating point: how far can an explicit,
inspectable extraction pipeline go without supervised training on the target
task, and what can its modular structure reveal about where extraction succeeds
or breaks down? We use \emph{target-task-unsupervised} to mean that no labeled
examples from the target extraction task are used for training or fine-tuning.
Pretrained resources may still be used; in particular, one candidate selector
uses a pretrained NLI model in a zero-shot configuration
~\cite{yin2019benchmarking}.

We present \posid{}, a modular framework that separates evidence localization,
provenance-controlled extraction scope, structured attribute recovery,
evidence traversal, and person association. Lexical, distributional,
lexical-semantic, and zero-shot NLI selectors can provide candidate evidence
to the same downstream extractor. Candidate provenance controls which
attribute operations are permitted, while an explicit traversal policy
determines whether clothing recovery terminates when local syntactic evidence
becomes unsupported or continues through the remainder of the candidate.
Keeping these stages explicit allows retrieval, traversal, and downstream
recovery behavior to be examined without silently changing the rest of the
pipeline.

\newpage
The contributions of this paper are:

\begin{enumerate}
    \item We introduce \posid{}, a modular person-attribute extraction
    framework that separates evidence localization, provenance-controlled
    extraction scope, structured recovery and traversal, and person
    association without supervised training on the target task. This design
    allows multiple candidate-localization mechanisms to feed the same
    explicit downstream recovery procedure.

    \item We introduce evidence-aware evaluation for collective and
    underspecified person descriptions. Provenance-aware clothing scoring
    distinguishes ordinary person-specific evidence from shared, distributed,
    and member-unspecified evidence, while a separate person-aware diagnostic
    measures ownership, fragmentation, and spurious-person errors. Together,
    these views separate evidence recovery from annotation cardinality and
    person attribution.

    \item We provide a controlled characterization of \posid{} across distinct
    document regimes while holding downstream recovery fixed. The experiments
    reveal a retrieval--traversal interaction: RE-containing configurations
    consistently benefit from conservative early stopping across the evaluated
    real-use cohorts, while standalone semantic retrieval can benefit from
    continuation when localization is less constrained. A verified-span
    intervention further shows that stronger candidate localization directly
    improves \posid{}, while the remaining gap identifies additional
    structured-recovery and person-attribution challenges.
\end{enumerate}


\section{The \incitext{} Dataset}
\label{sec:dataset}

\incitext{} is a person-attribute text resource assembled from public-safety
material and controlled text generation. We used \incitext{} v2~\cite{solaiman2026incitext} in this
study. It contains 572 released source-text documents: a 558-document
consolidated core and the 14-document Challenge14 extension. The core comprises
10 WLPD-provided Synthetic descriptions, 13 press releases, 5 incident/dispatch
records, 30 longer synthetic \texttt{Generated} descriptions, and 500
compact synthetic \texttt{Generated\_flash} descriptions. In this paper, we refer to the 30-document Generated subset as \textbf{Generated30} and the 500-document subset as \textbf{Compact500}. The first three source groups form \textbf{Base28}.

\textbf{Challenge14} contains 14 additional documents---13 incident/dispatch-style
records and one press release---with 26 annotated people; nine documents are
multi-person, one is single-person, and four contain no annotated target
person. \textbf{Facebook21} is a separate external cohort of 21 randomly selected public
missing-person notices from the \emph{Missing People in America} Facebook page.
The records were manually annotated and verified.
%
%
We preserve the cohort boundaries because they represent substantially different document and evidence conditions rather than interchangeable samples from one homogeneous corpus.

All documents in Base28 and Challenge14 were carefully curated and provided by the West Lafayette Police Department (WLPD) and manual descriptions were written by investigating officers familiar with report writing to capture different ways person attributes may be expressed.
\textit{Generated30} and \textit{Compact500} were created with synthetic event/address context. 
%

\begin{figure*}[h]
    \centering
    \includegraphics[
        width=\textwidth,
        trim={0 0 0 82},
        clip
    ]{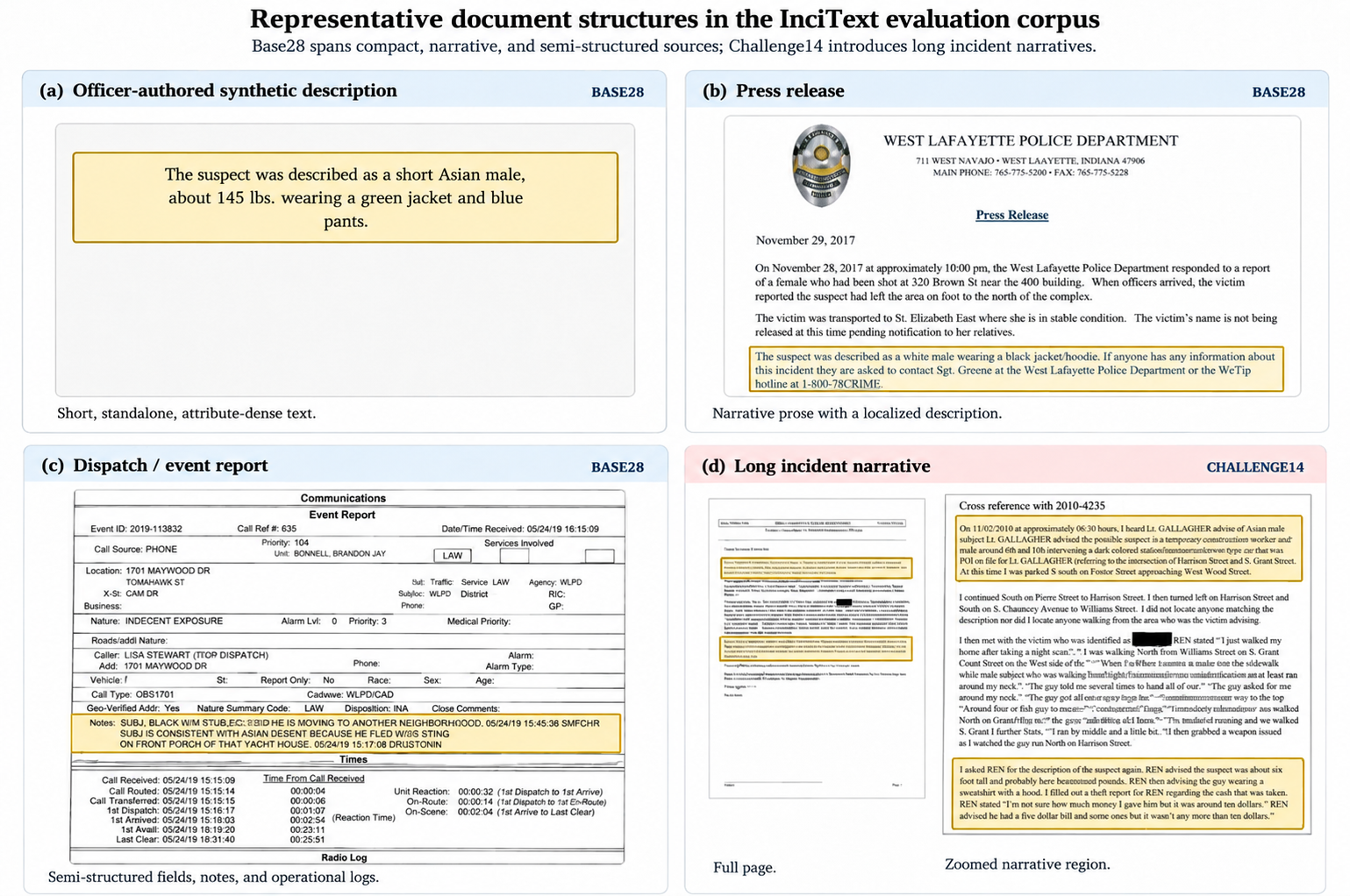}
    \caption{Document structures in the InciText evaluation corpus. Base28 spans compact attribute-dense descriptions, narrative press releases, and semi-structured dispatch/event reports. Challenge14 extends to long incident narratives in which person-attribute evidence can occur at multiple locations. Highlighted regions indicate representative person-description evidence.}
    \label{fig:source-heterogeneity}
\end{figure*}

Table~\ref{tab:dataset-cohorts} summarizes the five evaluation cohorts, while
Fig.~\ref{fig:source-heterogeneity} shows the principal document structures represented in Base28 and Challenge14.

\begin{table}[t]
\centering
\caption{Evaluation cohorts used in this study.}
\label{tab:dataset-cohorts}
\small
\begin{tabular}{lrrp{4.0cm}}
\toprule
Cohort & \#Docs & \#Person & Role \\
\midrule
Base28      & 28  & 23  & Mixed-source base cohort \\
Challenge14 & 14  & 26  & Heterogeneous multi-person challenge cohort \\
\midrule
Generated30 & 30  & 30  & Larger documents created with synthetic context \\
Compact500  & 500 & 500 & Compact description of person with synthetic context \\
\midrule
Facebook21  & 21  & 21  & External naturalistic robustness cohort \\
\bottomrule
\end{tabular}
\end{table}

\subsection{Person-Level Annotation and Evidence Provenance}
\label{subsec:dataset-annotation}

Annotations are organized around people rather than only around documents.
Each document identifier maps to zero or more person records, allowing
attributes belonging to different people to remain separate. The broader
annotation schema covers demographic, physical, and appearance information;
the present study evaluates \texttt{gender, race, height,} and \texttt{clothing}. Clothing is
represented through the nested \texttt{wearing} field, which associates each
item with zero or more textual descriptors. Thus, for example, 
a description containing a green jacket and blue pants is encoded as \texttt{jacket:[green]} and \texttt{pants:[blue]}.
%


Missing fields indicate that no corresponding value is annotated from the
source text; they are not interpreted as negative evidence. Challenge14
additionally records evidence provenance for cases in which 
person-cardinality and linguistic evidence do not align. The annotations distinguish ordinary person-specific evidence from shared or subgroup evidence.
Section~\ref{subsec:experimental-evaluation} defines how they are used during evaluation.


\section{\posid{} Method}
\label{sec:method}

\subsection{Problem Formulation and Overview}
\label{subsec:method-overview}

Let a document be an ordered sequence of sentences
$T=\langle s_1,\ldots,s_n\rangle$. Given a small set of domain-relevant query
phrases $Q$ and a target attribute set $O$, the task is to recover a
person-indexed collection of structured attribute--value pairs from $T$. In
this work,
\[
O=\{\textsc{gender},\textsc{race},\textsc{height},\textsc{clothing}\},
\]
where the first three attributes have constrained value forms and clothing is
open-vocabulary: both the item name and its descriptors may vary with the
text.

\posid{} separates three concerns that are often conflated in end-to-end
extraction: \emph{evidence localization}, \emph{structured recovery within the
localized evidence}, and \emph{association of recovered evidence with person
records}. Figure \ref{fig:posid-architecture} shows the full system architecture.

\begin{figure*}[t]
\centering
\begin{tikzpicture}[
    >=Latex,
    box/.style={
        draw=gray!65!black,
        fill=white,
        rounded corners=2pt,
        align=center,
        inner sep=5pt,
        minimum height=9mm,
        font=\scriptsize,
        line width=0.65pt
    },
    retrieval/.style={
        box, draw=blue!65!black, fill=blue!7
    },
    interface/.style={
        box, draw=blue!80!black, fill=blue!16,
        line width=0.85pt
    },
    recovery/.style={
        box, draw=teal!65!black, fill=teal!9
    },
    clothing/.style={
        box, draw=orange!75!black, fill=orange!20,
        line width=0.85pt
    },
    output/.style={
        box, draw=teal!85!black, fill=teal!22,
        line width=0.85pt
    },
    arrow/.style={
        ->, thick, draw=gray!80!black
    },
    branch/.style={
        thick, draw=gray!80!black
    },
    stage/.style={
        font=\scriptsize\bfseries,
        text=gray!80!black
    }
]

\node[retrieval] (doc) at (0,0)
    {Document $T$};
\node[retrieval] (retrieve) at (3.0,0)
    {Candidate retrieval\\RE / W2V / WN / NLI};
\node[interface] (cand) at (6.2,0)
    {\texttt{CandidateSentence}\\text, score, query, context};
\node[interface] (scope) at (9.7,0)
    {Provenance-controlled scope\\fixed / clothing / both};

\node[stage] at (4.8,0.85)
    {Stage 1: Evidence localization};

\node[recovery] (fixed) at (8.0,-2.0)
    {Fixed attributes\\gender, race, height};
\node[clothing] (cloth) at (11.5,-2.0)
    {Clothing parser\\stop / continue};
\node[recovery] (mentions) at (8.0,-3.8)
    {Attribute mentions};
\node[recovery] (resolve) at (11.5,-3.8)
    {Person resolver};
\node[output] (out) at (14.7,-3.8)
    {Person-indexed\\attributes};

\node[stage] at (12.5,-1.05)
    {Stage 2: Structured recovery and association};

\begin{scope}[on background layer]
    \filldraw[
        fill=blue!2,
        draw=blue!45!black,
        dashed,
        rounded corners=4pt,
        line width=0.7pt
    ] (-1.15,-0.65) rectangle (11.5,0.65);

    \filldraw[
        fill=teal!2,
        draw=teal!45!black,
        dashed,
        rounded corners=4pt,
        line width=0.7pt
    ] (6.75,-1.35) rectangle (16.0,-4.4);
\end{scope}

\draw[arrow] (doc) -- (retrieve);
\draw[arrow] (retrieve) -- (cand);
\draw[arrow] (cand) -- (scope);

\coordinate (junction) at (9.7,-1.38);
\draw[branch] (scope.south) -- (junction);
\fill[gray!80!black] (junction) circle (1.5pt);

\draw[arrow] (junction)--++(0,.2)-|(fixed.north);
\draw[arrow] (junction)--++(0,.2)-|(cloth.north);

\draw[arrow] (fixed.south) -- (mentions.north);
\draw[arrow] (cloth.south) -- ++(0,-0.50)
    -| ([xshift=8mm]mentions.north);
\draw[arrow] (mentions) -- (resolve);
\draw[arrow] (resolve) -- (out);

\end{tikzpicture}

\caption{\posid{} separates evidence localization from
structured recovery and person association. Candidate
provenance controls extraction scope; the clothing parser
supports conservative stopping and continued traversal.}


\label{fig:posid-architecture}
\end{figure*}

\subsection{Candidate Representation and Selectors}
\label{subsec:candidate-representation}

Every selector returns the same provenance-bearing representation
\[
c=\langle i,r,q,\kappa\rangle,      
\]
where $i$ is the sentence index, 
$r$ is its relevance score, $q$ is triggering query or pattern 
and $\kappa$ is the complete sentence or expanded local context. 
The triggering query determines which extraction roles are enabled downstream.  


For sentence fragments beginning with a form of \emph{wear} (e.g., ``wearing
a blue jacket''), the immediately preceding sentence is attached to $\kappa$
when it contains an explicit person-role anchor such as \emph{suspect},
\emph{victim}, or \emph{juvenile}. This handles a common fragmentation pattern
in which a person is introduced in one sentence and the clothing description begins in the next.


We compare four selectors, as shown in Table~\ref{tab:selectors}. RE emits sentences matching the literal cue \texttt{wearing} or a fixed suspect/demographic pattern. Word2Vec~\cite{mikolov2013efficient} scores a sentence by the maximum token--query cosine similarity. WordNet~\cite{wordnet98} uses noun-synset Wu--Palmer similarity. Zero-shot NLI~\cite{yin2019benchmarking} scores the label \texttt{clothes} with 
the explicit hypothesis template 
\[\texttt{This example is \{\}.}\] Each selector emits every qualifying sentence rather than stopping after the first semantic match.


\begin{table}[t]
\centering
\caption{Candidate selectors used by \posid{}. 
}
\label{tab:selectors}
\small
\begin{tabular}{p{2cm}
p{2.2cm}p{3.6cm}}
\toprule
Selector & 
Downstream Context &
Downstream Scope \\
\midrule
RE &
Matched \newline pattern & 
\texttt{wearing}: fixed + clothing; fixed pattern: fixed only \\

Word2Vec &
Full sentence &
Clothing only \\

WordNet &
Full sentence &
Clothing only \\

Zero-shot NLI &
Full sentence &
Clothing only \\
\bottomrule
\end{tabular}
\end{table}

Combined RE+WordNet and RE+NLI configurations concatenate the ordered RE candidates with semantic clothing candidates rather than using the semantic selector only as fallback. If $C_R$ denotes the ordered RE candidates and $C_M$ the semantic clothing candidates, the combined stream is
\begin{equation}
C=C_R\,\Vert\,C_M,
\end{equation}
where $\Vert$ denotes ordered concatenation. Candidate provenance prevents broader semantic retrieval from silently changing the fixed-attribute evidence path: \textbf{\texttt{wearing}} may support fixed attributes and clothing, the fixed RE pattern supports only fixed attributes, and semantic \textbf{\texttt{clothes}} candidates support only clothing. Because the same sentence may enter through multiple selectors or roles, downstream processing uses a role-aware key so one extraction scope does not suppress another.

\begin{algorithm}[t]
\caption{Combined candidate retrieval and structured \posid{} recovery}
\label{alg:posid}
\KwIn{document $T$; RE selector $R$; optional semantic selector $M$; parser mode $\tau$}
\KwOut{person-indexed structured attributes $\widehat{A}$}
$C_R\leftarrow R(T,\{q_{\mathrm{wear}},q_{\mathrm{fixed}}\})$\;
\eIf{$M$ is enabled}{$C_M\leftarrow M(T,\{\text{clothes}\})$\;}{$C_M\leftarrow[\ ]$\;}
$C\leftarrow C_R\,\Vert\,C_M$\;
construct person anchors and local/shared context\;
$L\leftarrow\varnothing$; $K\leftarrow\varnothing$\;
\ForEach{$c\in C$}{
  $(b_f,b_c)\leftarrow\operatorname{Role}(c.q)$\;
  $u\leftarrow\operatorname{Context}(c,T)$; $k\leftarrow(c.i,u,b_f,b_c)$\;
  \If{$k\in K$}{continue\;}
  $K\leftarrow K\cup\{k\}$\;
  \If{$b_f$}{
    $L_f\leftarrow\operatorname{Fixed}(u)$; $L\leftarrow L\cup L_f$\;
    \If{$L_f$ contains no height and next sentence exists}{$L\leftarrow L\cup\operatorname{HeightOnly}(s_{c.i+1})$\;}
  }
  \If{$b_c$}{$L\leftarrow L\cup\operatorname{ClothingParser}(u,\tau)$\;}
}
$\widehat{A}\leftarrow\operatorname{ResolvePeople}(L)$\;
\Return $\widehat{A}$\;
\end{algorithm}

\subsection{Fixed and Open-Vocabulary Attribute-value Recovery}
\label{subsec:fixed-open-extraction}
Fixed attributes use 
explicit lexical and contextual patterns.
Gender is recognized only from explicit whole-word forms and canonicalized to the target representation; no gender is inferred from names. Race is accepted only in local race--gender adjacency, in either order, which reduces collisions between demographic terms and clothing colors. Height accepts common numeric forms (feet/inches, hyphenated feet, and quote notation) and constrained qualitative constructions such as \emph{short Asian male}; generic uses such as \emph{short hair} are therefore excluded. If a fixed-attribute candidate contains no height, the immediately following sentence is inspected for height only, while person anchors continue to constrain ownership. 
The complete patterns and normalization rules are provided
in Appendix~\ref{app:fixed_patterns}.

Clothing is treated differently as open-vocabulary because the set of possible item names and
modifiers is not closed.
For a clothing-enabled candidate, \posid{}  begins parsing at a \emph{wear} cue when present; otherwise spans already identified as fixed attributes are removed and the remaining candidate is parsed.

The parser maintains a descriptor buffer $D$ and a current noun
run $N$ to recover clothing items and their descriptions.
Adjectives are accumulated in $D$, while consecutive nouns
are grouped into compound item names, such as \emph{tank top}
or \emph{dress pants}. Because part-of-speech tagging is
imperfect, noun tokens semantically associated with the
WordNet sense of \emph{color} are treated as descriptors
rather than clothing items. Contextual modifiers such as
\emph{dark} and \emph{light} are similarly retained as
descriptors when they modify a following noun.
The parser also preserves selected pre-nominal verb/particle
constructions, such as \textit{``buttoned up shirt,''} as descriptions
of the clothing item. Once an item boundary is reached,
the parser emits $\langle\textit{item},D\rangle$ and resets
the corresponding buffers. Clothing names retain their
surface forms rather than being mapped to a predefined
ontology. 
Slash-separated item expressions are split into
individual records before output.

Unexpected part-of-speech (POS) patterns create an explicit traversal choice. Under \emph{early stop}, the first otherwise unsupported POS pattern 
terminates scanning. Under \emph{continue}, the current item description is considered complete but scanning resumes, allowing later clothing evidence to be found at the cost of potential false positives. Crucially, this choice is downstream of retrieval: the same localized candidate can be processed under either traversal policy.
%
The complete clothing-parser pseudocode is given in Appendix~\ref{app:algorithm}.

\subsection{Person Association}
\label{subsec:person-resolution}
Attribute extraction and person assignment are separated. Before candidate processing, the document is scanned for narrow person anchors, including explicit roles (e.g., suspect, victim, juvenile) and name-like tokens after demographic/role stopword filtering. Extracted mentions prefer the nearest preceding explicit anchor in local context. Pronouns and recent active records provide fallback context; already extracted gender may constrain \emph{he}/\emph{she} association but never creates a new gender value. Coordinated anchors followed by \emph{both} can share a clothing mention. Otherwise, unanchored evidence goes to the nearest active person or a deterministic anonymous record. This resolver is intentionally lightweight and imperfect, but keeping it explicit allows document-level recovery and person-level ownership failures to be measured separately.

\section{Experimental Settings}
\label{sec:experiments}


We evaluate six candidate-localization configurations---RE, Word2Vec,
WordNet, zero-shot NLI, RE+WordNet, and RE+NLI -- under the two traversal
policies:
\emph{early stop} and
\emph{continue}. Base28, Challenge14, and Facebook21 form the main real-use comparison, while
Generated30 and Compact500 are used separately as synthetic
diagnostics. Challenge14 additionally receives two standalone diagnostics:
a direct generative baseline and an oracle localization intervention.




The Word2Vec, WordNet, and NLI selectors use thresholds of $0.50$, $0.90$,
and $0.85$, respectively. Word2Vec uses the NLTK
\texttt{word2vec\_sample} vectors, WordNet uses noun-synset Wu--Palmer
similarity, and NLI uses \texttt{facebook/bart-large-mnli}. RE uses the
\texttt{wearing} cue and the fixed demographic pattern described by the
method. 
Apart from the candidate source and traversal policy, extraction,
normalization, and person association are held fixed.


The generative baseline uses GPT-5.6 Sol in a zero-shot, high-reasoning
configuration, with one independent stateless request per document. Each
request contains only the frozen task instructions, structured-output schema,
and source document; no tools, retrieval, conversation history, POSID output,
evaluator feedback, or gold annotations are provided. No LLM-specific
canonicalization or post-hoc repair is applied before scoring.

The oracle condition replaces automatic clothing localization with manually
verified minimum source span(s) supporting the annotated clothing evidence.
No gold person identifiers, roles, or attribute values are supplied to
\posid{}. All downstream parsing, traversal, normalization, person association,
and evaluation remain unchanged. The same oracle evidence is evaluated under
both traversal policies; it is therefore a localization diagnostic rather than
an upper bound on \posid{}.

\subsection{Evaluation Metrics}
\label{subsec:experimental-evaluation}

We report micro-aggregated precision, recall, and $F_1$. Fixed attributes are
matched as normalized document-level multisets. Clothing is evaluated as
Attr-Only (AO), which scores item recovery, and Attr-Value (AV), which also scores associated descriptors and gives partial credit when an item is correctly recovered but its description is incomplete.

A separate person-aware diagnostic scores recovered evidence within matched-person records, exposing ownership, fragmentation, and spurious-person errors that document-level AO/AV can hide. We additionally use provenance-aware clothing scoring for source statements whose ownership is collective or underspecified. 
Explicitly tagged shared evidence contributes a single recovery obligation; distributed group evidence preserves the stated clothing multiset while allowing ownership within the tagged group to remain exchangeable; and member-unspecified evidence may be satisfied by any member of the tagged group. Untagged person-specific evidence retains ordinary scoring. 
%
%


\section{Results and Analysis}
\label{sec:results}

\subsection{Recovery Across Document Conditions}
\label{subsec:results-regimes}

Table~\ref{tab:regime-summary} summarizes the strongest clothing recovery results observed under different configurations for \posid{}. 
The contrasts describe the evaluated conditions rather than isolating document heterogeneity: the cohorts differ in source, structure, length, and person composition. 
On the two comparatively clean and shorter cohorts, 
\posid{} operates at or near ceiling: 
Compact500 reaches 1.000 AO and .972 AV, while Generated30 reaches 1.000 on both metrics 
with the RE-containing methods under continuation. \posid{} performance remains comparatively strong on the moderate-length real-world public-safety cohorts, 
with Base28 reaching .824/.857 AO/AV and Facebook21 .667/.651. The largest degradation appears on Challenge14, where the best observed AO and AV fall to .500 and .382, respectively. Challenge14 contains substantially longer, heterogeneous, multi-person narratives with shared or underspecified descriptions and less regular attribute-expression patterns. These characteristics plausibly contribute to the degradation by making relevant evidence harder to localize, increasing the amount of irrelevant material encountered during traversal, and complicating person-level association. 
A stronger zero-shot generative baseline improves recovery on Challenge14 while remaining below ceiling, further indicating that the cohort is difficult, shown in Table~\ref{tab:recovery-diagnostic}.
\textit{We also observe that RE+NLI is consistently among the strongest configurations across cohorts}: tied on Compact500, Generated30, Facebook21, and Challenge14 AV F1, and remains within .010 AO and .005 AV of the best Base28 configuration, RE+WordNet.
Overall, these results suggest that \posid{} is highly effective when attribute evidence is cleanly expressed, remains useful on moderate naturalistic descriptions, and degrades as descriptions become longer, less regularly structured, and more ownership-ambiguous.

\begin{table*}[t]
\centering
\caption{Best observed clothing $F_1$ by cohort. 
Only Challenge14's best AO and AV comes from different configurations.
}
\label{tab:regime-summary}
\small
\setlength{\tabcolsep}{6pt}
\begin{tabular}{lccll}
\toprule
Cohort & Best AO & Best AV & Best Configuration(s) & Traversal Policy \\ 
\midrule
Base28 & .824 & .857 & \textbf{RE+WordNet} & early stop \\
Challenge14 & .500 & .382 & AO: NLI; AV: \textbf{RE+NLI} & AO: continue; AV: early stop \\
Facebook21 & .667 & .651 & RE, \textbf{RE+NLI} & early stop \\
\midrule
Compact500 & 1.000 & .972 & RE, NLI, RE+WordNet, \textbf{RE+NLI} & continue \\
Generated30 & 1.000 & 1.000 & RE, RE+WordNet, \textbf{RE+NLI} & both identical \\
\bottomrule
\end{tabular}
\end{table*}

The difference extends beyond clothing. Along the common RE fixed-attribute path, gender/race/height \(F_1\) changes from .789/.811/.818 on Base28 to .323/.211/.000 on Challenge14. Because the semantic selectors augment clothing localization only, this loss is not remedied by adding semantic clothing candidates.
The degradation is consistent with the fixed attributes relying on regex-based pattern matching, while real-world descriptions vary across platforms, for example, police reports and Facebook missing-person posts---and often express these attributes in different free-form patterns. 

\subsection{Traversal Policy Across Real-Use Conditions}
\label{subsec:traversal-policy}

Table~\ref{tab:traversal-delta} compares early stopping and continuation on the three actual use-case cohorts. Across Base28, Challenge14, and Facebook21, every RE-containing configuration performs better with early stopping. The pattern is less uniform for the standalone semantic selectors. On Base28, NLI favors early stopping while Word2Vec and WordNet favor continuation; on Challenge14, all three semantic selectors favor continuation; and on Facebook21, all selectors favor early stopping. The Facebook21 results further show that the benefit of early stopping is primarily a precision effect: for RE, AO false positives decrease from 19 to 8 while true positives change only from 16 to 15. Thus, continued traversal can recover additional evidence when semantic localization is incomplete, but can also admit unrelated material when candidate boundaries are less precise. Early stopping instead provides a consistent safeguard for the RE-based pipelines, whose initial localization already supplies a more constrained candidate region. Together with the cross-cohort results in Table~\ref{tab:regime-summary}, this makes \textit{RE+NLI with early stopping the most stable \posid{} configuration across the evaluated real-use conditions}, although standalone semantic retrieval can benefit from continuation on the harder narratives.

\begin{table}[t]
\centering
\caption{Continuation effect on clothing $F_1$ ($\Delta=\mathrm{continue}-\mathrm{early}$). Positive favors continuation; negative favors early stopping. Facebook21 deltas use full-precision primary counts.}
\label{tab:traversal-delta}
\footnotesize
\setlength{\tabcolsep}{5.0pt}
\begin{tabular}{lrr|rr|rr}
\toprule
 & \multicolumn{2}{c}{Base28} & \multicolumn{2}{c}{Challenge14} & \multicolumn{2}{c}{Facebook21} \\
\cmidrule(lr){2-3}\cmidrule(lr){4-5}\cmidrule(lr){6-7}
Selector & AO & AV & AO & AV & AO & AV \\
\midrule
Word2Vec   & +.006 & +.009 & +.100 & +.101 & -.068 & -.053 \\
WordNet    & +.012 & +.016 & +.142 & +.119 & -.056 & -.043 \\
NLI        & -.022 & -.009 & +.050 & +.081 & -.086 & -.044 \\
\midrule
RE         & -.034 & -.017 & -.107 & -.044 & -.105 & -.063 \\
RE+WordNet & -.037 & -.018 & -.102 & -.042 & -.110 & -.063 \\
RE+NLI     & -.067 & -.035 & -.134 & -.059 & -.105 & -.063 \\
\bottomrule
\end{tabular}
\end{table}


\subsection{Better Candidate Localization Improves POSID}
\label{subsec:results-localization}

Table~\ref{tab:recovery-diagnostic} evaluates whether the Challenge14 degradation can be reduced by improving candidate localization while leaving the remainder of \posid{} unchanged. Under the early-stopping policy identified in the preceding analysis, replacing automatic RE+NLI candidates with manually verified clothing spans improves both clothing metrics: 
AO \(F_1\) increases by 8.8\%, and AV \(F_1\) by 7.3\%.
The improvement is primarily precision-driven: AO false positives decrease from 16 to 12 while true positives and false negatives remain unchanged at 12 and 10. Thus, higher-quality candidate localization directly improves \posid{} without modifying its parser, normalization, or person-association components.

Continuation provides a complementary diagnostic. With verified spans, AO true positives increase from 13 to 18 and false negatives decrease from 9 to 4, showing that additional relevant evidence becomes accessible; however, the broader traversal also increases false positives from 17 to 30. The zero-shot generative reference reaches .833 AO and .735 AV on the same cohort, indicating that 
additional gains are possible from different modules.
Together, these results show that \posid{}'s modular design enables recovery improvements through stronger candidate localization.

\begin{table*}[t]
\centering
\caption{Challenge14 diagnostic results under identical clothing evaluation. AO TP/FP/FN counts use 22 provenance-adjusted gold clothing-evidence obligations. 
C: continue; E: early stop.}
\label{tab:recovery-diagnostic}
\small
\setlength{\tabcolsep}{5pt}
\begin{tabular}{lccccccc}
\toprule
\makecell{Candidate source / \\ extractor / policy} & \makecell{AO TP \\ (Correct items)} & \makecell{AO FP \\ (Unsupported items)} & \makecell{AO FN \\ (Missed items)} & AO $F_1$ & AV $F_1$ & Person AO & Person AV \\
\midrule
NLI / \posid{} / C  & 13 & 17 & 9 & .500 & .378 & \textit{.385} & \textit{.230} \\
Verified spans / \posid{} / C & 18 & 30 & 4 & .514 & \textit{.429} & .314 & .228 \\
RE+NLI / \posid{} / E & 12 & 16 & 10 & .480 & .382 & .160 & .103 \\
Verified spans / \posid{} / E  & 12 & 12 & 10 & \textit{.522} & .410 & .261 & .148 \\
GPT-5.6 Sol (zero-shot) / -- & 20 & 6 & 2 & \textbf{.833} & \textbf{.735} & \textbf{.792} & \textbf{.613} \\
\bottomrule
\end{tabular}
\end{table*}

\subsection{Evidence Recovery is separate from Person Attribution} 
\label{subsec:results-ownership}

Lightweight person-aware scoring exposes a separate limitation that document-level AO/AV cannot measure. Generated30 illustrates the distinction without a difficult retrieval setting: RE-containing configurations reach 1.000/1.000 primary clothing $F_1$ but approximately .861/.861 person-aware $F_1$, with 35 predicted person records for 30 annotated people. 
On Challenge14, NLI with continuation falls from .500/.378 primary to .385/.230 person-aware, and RE+NLI with early stopping falls from .480/.382 to .160/.103. 
Even the generative reference drops from .833/.735 primary to .792/.613 person-aware  in Table~\ref{tab:recovery-diagnostic}. 

As a regression check, provenance-aware scoring changes Challenge14 RE+NLI early-stop AO/AV \(F_1\) from .429/.337 to .480/.382 while the predictions remain identical; Base28, which contains no tagged collective-evidence cases, is unchanged.

\section{Discussion, Limitations, and Future Work}
\label{sec:discussion}

The experiments separate three issues that an end-to-end extraction score can
otherwise conflate: recovering relevant evidence, structuring that evidence,
and assigning it to the appropriate person. Provenance-aware scoring addresses
the complementary evaluation problem that collective or underspecified source
statements can support fewer evidence obligations than the number of stored
person records. Special handling is therefore restricted to explicitly tagged
cases, while ordinary person-specific evidence retains standard scoring.
Person-aware evaluation separately measures ownership and fragmentation.

The evaluated attribute inventory is intentionally narrow. Gender,
race, height, and clothing have enough coverage for the present experiments,
while other annotated fields are sparse. 
Finally, ``target-task-unsupervised'' refers specifically to the absence of supervised training or fine-tuning on labeled \incitext{} examples from the target extraction task; pretrained models are still used for NLI and the generative reference.

Several extensions follow directly from the observed error patterns. Person
association is currently handled with lightweight rules and could be strengthened
with more capable mention- or entity-resolution methods. Semantic localization
is applied primarily to clothing, while gender, race, and height remain on the
fixed-attribute path; analogous contextual recovery mechanisms may therefore
help with less regular expressions of these attributes. The clothing oracle is
likewise a focused localization diagnostic rather than an equivalent diagnostic
for the fixed attributes.
The present study evaluates extraction quality rather than computational
efficiency. A direct comparison of latency, memory use, token usage, monetary
cost, and deployment requirements between \posid{} and generative extraction
remains future work, as does broader evaluation across additional agencies,
platforms, writing styles, and attribute types.



The system extracts attributes explicitly stated in the source text and does not
establish that those statements are factually correct or complete. Demographic
categories also simplify real-world identity and can reflect source-reporting,
annotation, lexical, and missingness biases. The resulting outputs should
therefore be interpreted as structured recovery of textual claims rather than
inference of unobserved personal characteristics.

\section{Related Work}
\label{sec:related}

Public-safety information extraction has long used lexical resources,
hand-crafted rules, and statistical NLP to convert narrative reports into
structured information. Iriberri and Leroy~\cite{iriberri2007crime} extract
standardized suspect-description attributes from witness narratives using
lexical resources and hand-crafted rules, while Ku et
al.~\cite{ku2008crimeie} extract people, clothing, vehicles, weapons, scenes,
and temporal information from police and witness narratives using
crime-specific lexicons and rule-based extraction modules. Later work
emphasizes the operational value of unstructured police text and the practical
difficulty of analyzing it at scale~\cite{dixon2021policing}, while specialized
criminal-text systems have also combined domain lexicons, hand-crafted rules,
and learned components when large labeled resources are
unavailable~\cite{skorzewski2022criminal}. A recent scoping review documents
the breadth of computational analysis of unstructured police data, including
the growing use of pretrained language models~\cite{lukmanjaya2026police}. 
\posid{} operates in this problem family, but formulates extraction as explicit stages rather than as a single end-to-end task.

Contemporary public-security extraction includes LLM-based extraction from
police reports using prompting, augmentation, and LoRA
fine-tuning~\cite{xing2024police}, and TIPS, which combines expert-informed
synthetic data, domain schemas, and fine-tuned language models for
public-security information extraction~\cite{liu2025tips}. In the
missing-person domain, the Guardian Parser Pack uses source-specific
deterministic parsers and optional LLM-assisted extraction with schema
harmonization and validation~\cite{castillo2026parser}; a related Guardian
system employs task-specialized LLMs, consensus, and QLoRA
adaptation~\cite{castillo2026guardian}. These studies demonstrate the scope
of learned, generative, and schema-oriented approaches in the application
domain. \posid{} examines a complementary setting: extraction without
supervised training or fine-tuning on the target task, with explicit stages
whose behavior can be examined separately.

A broader line of work addresses information extraction with reduced
task-specific supervision. Zero-shot slot filling can be formulated as
question answering over slot descriptions~\cite{du2021qasf}, and relation
extraction as entailment over verbalized relation
labels~\cite{sainz2021entailment}. UIE~\cite{lu2022uie} instead learns a
unified text-to-structure generator across several IE tasks and schemas, while GoLLIE~\cite{sainz2024gollie} improves transfer to unseen extraction tasks by
training models to follow annotation guidelines. GLiNER~\cite{zaratiana2024gliner} demonstrates compact, open-type
named-entity recognition with zero-shot capabilities for resource-constrained
settings. Recent LLM-based IE work shows that language models can directly generate structured extraction outputs~\cite{shamsabadi2024llmie}, while adherence to the required output structure can itself constitute a distinct challenge~\cite{li2024structured}. In contrast, \posid{} uses lexical and
pretrained semantic resources for candidate localization while retaining an
explicit procedure for structured attribute recovery and person association;
the generative baseline serves as a separate reference, not a component of
\posid{}.

Our evaluation problem intersects work on entity attributes and non-singular
reference. Choi et al.~\cite{choi2015partial} study entity-attribute extraction from complex noun phrases paired with one or more entities, assigning supported attributes to the described entities.
More generally, plural mentions violate the
usual one-mention/one-entity assumption: Zhou and Choi~\cite{zhou2018plural} introduce plural mentions
for coreference resolution and entity linking and modify evaluation to account
for their set-valued semantics.  Split-antecedent work~\cite{yu2021split, yu2022anaphora} likewise shows that jointly evaluating single- and split-antecedent anaphora requires extending or carefully adapting standard coreference evaluation. Our setting is adjacent but different. Rather than resolving general plural coreference, we
examine how many attribute-recovery obligations are supported by a source
statement and whether ownership is individual, shared, distributed, or
member-unspecified. Provenance-aware clothing scoring therefore distinguishes individually supported,
shared, distributed, and member-unspecified clothing evidence while preserving
ordinary person-specific scoring elsewhere, and person-aware evaluation
separately measures ownership and fragmentation.



\section{Conclusion}
\label{sec:conclusion}

We presented \posid{}, a modular, target-task-unsupervised framework for
person-attribute extraction that separates evidence localization,
provenance-controlled extraction scope, structured recovery and traversal, and
person association. 
The evaluation shows that these stages interact rather
than contributing a single uniform source of error: traversal behavior depends
on how candidate evidence is localized, and replacing automatic candidates
with verified spans directly improves recovery while leaving additional
structured-recovery and person-attribution errors. Provenance-aware and
person-aware evaluation further separate evidence recovery from annotation
cardinality and ownership, making these failure modes observable rather than
collapsing them into a single end-to-end score.

\posid{} remains highly effective when attribute
evidence is sufficiently regular, while harder heterogeneous and multi-person
narratives expose distinct points at which localization, traversal, recovery,
and attribution can be improved. 
Across the evaluated regimes, \posid{} reaches near-ceiling
recovery on structurally regular descriptions and reveals a consistent operating
principle on real-use text: precise candidate localization paired with
conservative traversal provides the most reliable lightweight extraction, while
semantic continuation can be useful when localization is less constrained.
The strongest configurations reach 1.000/.972 AO/AV $F_1$ on Compact500,
while the best automatic Challenge14 AO/AV results are .500/.382.
This explicit decomposition preserves the
original lightweight, inspectable, on-demand operating point motivated by
public-safety workflows while providing a practical basis for diagnosing and
improving structured person-attribute extraction without supervised training
on the target task.

\section*{Acknowledgment}
This work began during the author's doctoral studies at Purdue University under
the supervision of Bharat Bhargava and was supported in its early stages by the
Northrop Grumman Mission Systems Research in Applications for Learning Machines
(REALM) Program. The author gratefully acknowledges the West Lafayette Police
Department for assistance with the original problem formulation, data
collection, and domain feedback, particularly Sergeant Troy Greene for his
support in arranging interviews and preparing incident-report material. The
author also thanks Jim MacDonald for feedback during the early development of
the project, Tomas Hrdlovics at Purdue University for assistance with early
implementation, Harshit Singh at Purdue University for assistance with data
annotation, and Aklank Jain at UMBC for assistance with annotation of the
Facebook missing-person cohort.

\balance

\clearpage
\appendices


\section{Attribute Coverage in \incitext}
\label{app:attribute-coverage}

Table~\ref{tab:attribute-coverage} reports the Base28 coverage of the
four attribute groups evaluated by POSID. These attributes provide sufficient
annotation support for the present analysis; other fields in the broader
annotation schema are substantially sparser.

\begin{table}[!htbp]
    \centering
    \caption{Gold annotation coverage for the attribute groups evaluated in
    Base28. Percentages use the 23 person-level annotation records as the
    denominator.}
    \label{tab:attribute-coverage}
    \small
    \begin{tabular}{lrr}
        \toprule
        Attribute group & Person records & Coverage \\
        \midrule
        Gender & 23 & 100.0\% \\
        Race & 22 & 95.7\% \\
        Height & 13 & 56.5\% \\
        Clothing (\texttt{wearing}) & 18 & 78.3\% \\
        \bottomrule
    \end{tabular}
\end{table}

\FloatBarrier


\section{Fixed-Attribute Patterns and Normalization}
\label{app:fixed_patterns}

Table~\ref{tab:fixed_patterns} summarizes the finite-valued
attribute rules used by POSID. Matching is case-insensitive and uses
word boundaries where applicable. Here, $N$ denotes a numeric component,
$R$ a race term, and $G$ a gender token.

\begin{table}[!htbp]
\centering
\caption{Fixed-attribute extraction and normalization rules.}
\label{tab:fixed_patterns}
\footnotesize
\setlength{\tabcolsep}{3pt}
\renewcommand{\arraystretch}{1.15}
\begin{tabular}{p{0.22\columnwidth}p{0.70\columnwidth}}
\toprule
Attribute & Rule and normalization \\
\midrule

Gender &
Match explicit whole-word forms
\texttt{male}, \texttt{man}, \texttt{boy},
\texttt{female}, \texttt{woman}, and \texttt{girl};
canonicalize to \texttt{male} or \texttt{female}. \\

\addlinespace

Race &
Accept an explicit race term only in local race--gender adjacency,
$R\,G$ or $G\,R$.
Retain the explicit race value. \\

\addlinespace

Height (numeric) &
Accept common numeric forms including
\texttt{N feet [N inches] [tall]},
\texttt{N-foot}, \texttt{N ft}, and quote notation
(e.g., \texttt{5'10"}).
Preserve the matched expression. \\

\addlinespace

Height (relative) &
Accept \texttt{short} or \texttt{tall} only in a local demographic
construction of the form
\texttt{(short|tall) [R] G};
preserve the qualitative value. \\

\bottomrule
\end{tabular}
\end{table}

If a fixed-attribute candidate contains no height, POSID additionally
examines the immediately following sentence for height, subject to the
same person-context constraints. Demographic attributes are never inferred
from a person's name.

\FloatBarrier


\section{Open-Vocabulary Clothing Parser}
\label{app:algorithm}

Algorithm~\ref{alg:clothing} gives the clothing-recovery procedure used
after candidate localization. The traversal parameter $\tau$ isolates the
difference between conservative early stopping and continued scanning.

\begin{algorithm}[h]
\caption{Open-vocabulary clothing parser}
\label{alg:clothing}
\KwIn{candidate context $u$; parser mode
$\tau\in\{\textsc{continue},\textsc{stop}\}$}
\KwOut{clothing facts $S=\{\langle item,D\rangle\}$}

$x\leftarrow$ wear-triggered span if present, otherwise context with
fixed-attribute spans removed\;

POS-tag $x$; initialize
$S\leftarrow\varnothing$,
descriptor buffer $D$,
and noun run $N$\;

\ForEach{token $(w,t)$ in order}{

    \If{$w$ is a boundary, conjunction, preposition, or punctuation}{
        flush current item\;
        continue\;
    }

    \If{$t$ is an adjective or selected contextual modifier}{
        append $w$ to $D$\;
        continue\;
    }

    \If{$t$ is a noun}{

        \If{$w$ is a person/scaffold token}{
            flush current item\;
            continue\;
        }

        \If{$w$ is WordNet-close to color}{
            flush current item\;
            append $w$ to $D$\;
            continue\;
        }

        append $w$ to noun run $N$\;
        continue\;
    }

    \If{$\tau=\textsc{stop}$ and parsing has begun}{
        break\;
    }

    flush current item\;
}

flush final item\;
split slash-separated item expressions\;
\Return $S$\;
\end{algorithm}

\FloatBarrier


\section{Provenance-Aware Collective-Evidence Scoring}
\label{subsec:collective-evidence-metric}

Challenge14 contains cases in which the number of stored person records does
not equal the number of clothing assertions supported by the source text.
For example, ``five males wearing hooded sweatshirts'' expresses one
collective clothing fact even if the annotation representation stores the
same sweatshirt under several people. Treating those copies as independent
gold obligations would confound evidence recovery with person cardinality.

The evaluator therefore applies special handling only to explicitly tagged
collective evidence. Table~\ref{tab:collective-provenance} summarizes the
four scoring semantics. Provenance handling is scoped to the tagged field and
group: an unrelated person with an identical clothing fact remains a separate
obligation, and unsupported descriptors remain false positives under AV.
Untagged person-specific evidence retains ordinary scoring.

\begin{table*}[!t]
    \centering
    \caption{Provenance semantics used for clothing evaluation. Special
    handling applies only to the explicitly tagged field and group.}
    \label{tab:collective-provenance}
    \small
    \setlength{\tabcolsep}{5pt}
    \renewcommand{\arraystretch}{1.12}
    \begin{tabular}{
        p{0.20\textwidth}
        p{0.33\textwidth}
        p{0.38\textwidth}}
        \toprule
        Provenance type &
        Source-supported evidence &
        Evaluation obligation \\
        \midrule

        Untagged &
        Clothing fact is supported for a specific annotated person. &
        Retain the ordinary instance; person-aware scoring requires the
        matched person. \\

        \addlinespace

        \texttt{shared\_identical} /
        \texttt{shared\_subgroup} &
        The same clothing fact is asserted collectively for a tagged
        group or subgroup. &
        Count one evidence obligation for that tagged shared fact;
        repeated stored or predicted realizations do not create artificial
        cardinality errors. \\

        \addlinespace

        \texttt{group\_distributed} &
        Several distinct clothing facts are supported for a group, but
        member-specific ownership is not stated. &
        Preserve the clothing multiset while making ownership exchangeable
        within the tagged group. \\

        \addlinespace

        \texttt{one\_of\_group\_unspecified} &
        One clothing fact belongs to an unspecified member of a tagged
        group. &
        Count one evidence obligation and allow any member of the tagged
        group to satisfy it. \\

        \bottomrule
    \end{tabular}
\end{table*}

\FloatBarrier


\section{Full Factorial and Auxiliary Results}
\label{app:results-full}

This appendix reports the complete selector--traversal results underlying the
comparisons summarized in the main text.


\begin{table}[!htbp]
\centering
\caption{Main-cohort clothing $F_1$ for all six localization configurations
and both traversal policies (E: early stop; C: continuation).}
\label{tab:full-base-challenge}
\scriptsize
\setlength{\tabcolsep}{2.5pt}
\begin{tabular}{lrrrrrrrr}
\toprule
& \multicolumn{4}{c}{Base28}
& \multicolumn{4}{c}{Challenge14} \\
\cmidrule(lr){2-5}
\cmidrule(lr){6-9}
Selector &
AO-E & AO-C & AV-E & AV-C &
AO-E & AO-C & AV-E & AV-C \\
\midrule
RE          & .795 & .761 & .830 & .813 & .444 & .337 & .366 & .322 \\
Word2Vec    & .529 & .535 & .540 & .549 & .200 & .300 & .105 & .206 \\
WordNet     & .414 & .426 & .400 & .416 & .222 & .364 & .111 & .230 \\
NLI         & .741 & .719 & .768 & .759 & .450 & .500 & .297 & .378 \\
RE+WordNet  & .824 & .787 & .857 & .839 & .435 & .333 & .361 & .319 \\
RE+NLI      & .814 & .747 & .852 & .817 & .480 & .346 & .382 & .323 \\
\bottomrule
\end{tabular}
\end{table}


\begin{table}[!htbp]
\centering
\caption{Facebook21 primary clothing extraction under continuation (C)
and early stopping (E). The cohort contains 22 AO gold items and
44 AV evaluation units.}
\label{tab:full-facebook21}
\scriptsize
\setlength{\tabcolsep}{3pt}
\begin{tabular}{llrrrrrr}
\toprule
Selector & Policy &
AO P & AO R & AO $F_1$ &
AV P & AV R & AV $F_1$ \\
\midrule

RE
& C & .457 & .727 & .561 & .517 & .682 & .588 \\
& E & .652 & .682 & .667 & .667 & .636 & .651 \\

\midrule

Word2Vec
& C & .529 & .409 & .462 & .531 & .386 & .447 \\
& E & .750 & .409 & .529 & .708 & .386 & .500 \\

\midrule

WordNet
& C & .357 & .227 & .278 & .400 & .227 & .290 \\
& E & .625 & .227 & .333 & .625 & .227 & .333 \\

\midrule

NLI
& C & .438 & .636 & .519 & .500 & .591 & .542 \\
& E & .619 & .591 & .605 & .632 & .545 & .585 \\

\midrule

RE+WordNet
& C & .432 & .727 & .542 & .484 & .682 & .566 \\
& E & .625 & .682 & .652 & .622 & .636 & .629 \\

\midrule

RE+NLI
& C & .457 & .727 & .561 & .517 & .682 & .588 \\
& E & .652 & .682 & .667 & .667 & .636 & .651 \\

\bottomrule
\end{tabular}
\end{table}


\begin{table}[!htbp]
\centering
\caption{Facebook21 false-positive reduction for representative selectors.
Counts are from the primary clothing evaluation.}
\label{tab:facebook21-counts}
\small
\setlength{\tabcolsep}{4pt}
\begin{tabular}{llrrr}
\toprule
Selector & Metric & Continue & Early stop & Change \\
\midrule

RE
& AO TP & 16 & 15 & -1 \\
& AO FP & 19 &  8 & -11 \\
& AV TP & 30 & 28 & -2 \\
& AV FP & 28 & 14 & -14 \\

\midrule

Word2Vec
& AO TP &  9 &  9 & 0 \\
& AO FP &  8 &  3 & -5 \\
& AV TP & 17 & 17 & 0 \\
& AV FP & 15 &  7 & -8 \\

\midrule

WordNet
& AO TP &  5 &  5 & 0 \\
& AO FP &  9 &  3 & -6 \\
& AV TP & 10 & 10 & 0 \\
& AV FP & 15 &  6 & -9 \\

\bottomrule
\end{tabular}
\end{table}


\begin{table}[!htbp]
\centering
\caption{Synthetic-diagnostic clothing $F_1$. Generated30 predictions are
identical under early stopping and continuation; Compact500 is reported
separately for both policies.}
\label{tab:full-generated}
\scriptsize
\setlength{\tabcolsep}{4pt}
\begin{tabular}{lrrrrrr}
\toprule
& \multicolumn{2}{c}{Generated30 (E=C)}
& \multicolumn{4}{c}{Compact500} \\
\cmidrule(lr){2-3}
\cmidrule(lr){4-7}
Selector &
AO & AV &
AO-E & AO-C & AV-E & AV-C \\
\midrule
RE         & 1.000 & 1.000 & .908 & 1.000 & .908 & .972 \\
Word2Vec   & .654  & .654  & .712 & .795  & .712 & .767 \\
WordNet    & .326  & .326  & .418 & .491  & .418 & .467 \\
NLI        & .901  & .901  & .908 & 1.000 & .908 & .972 \\
RE+WordNet & 1.000 & 1.000 & .908 & 1.000 & .908 & .972 \\
RE+NLI     & 1.000 & 1.000 & .908 & 1.000 & .908 & .972 \\
\bottomrule
\end{tabular}
\end{table}


\begin{table}[!htbp]
\centering
\caption{Fixed-attribute micro precision, recall, and $F_1$ for the common
RE-containing path.}
\label{tab:full-fixed}
\small
\begin{tabular}{llrrr}
\toprule
Cohort & Attribute & P & R & $F_1$ \\
\midrule

Base28
& Gender & 1.000 & .652 & .789 \\
& Race   & 1.000 & .682 & .811 \\
& Height & 1.000 & .692 & .818 \\

\addlinespace

Challenge14
& Gender & 1.000 & .192 & .323 \\
& Race   & 1.000 & .118 & .211 \\
& Height & .000  & .000 & .000 \\

\bottomrule
\end{tabular}
\end{table}

\FloatBarrier

\end{document}